\documentclass[10pt,twocolumn,letterpaper]{article}
\usepackage{algorithm}
\usepackage{algcompatible}
\usepackage{enumitem}
\usepackage{bm}
\usepackage{cvpr}              

\usepackage[dvipsnames]{xcolor}

\definecolor{cvprblue}{rgb}{0.21,0.49,0.74}
\usepackage[pagebackref,breaklinks,colorlinks,citecolor=cvprblue]{hyperref}

\def\paperID{*****} 
\def\confName{CVPR}
\def\confYear{2024}

\title{LLM-AD: Large Language Model based Audio Description System}

\author{
Peng Chu \quad Jiang Wang \quad Andre Abrantes\\
Microsoft \\
{\tt\small \{pengchu, jiangwang, abrantes\}@microsoft.com}
}

\makeatletter
\newcommand{\thickhline}{%
    \noalign {\ifnum 0=`}\fi \hrule height 1pt
    \futurelet \reserved@a \@xhline
}
\makeatother

\begin{document}
\maketitle
\begin{abstract}

The development of Audio Description (AD) has been a pivotal step forward in making video content more accessible and inclusive. 
Traditionally, AD production has demanded a considerable amount of skilled labor, while existing automated approaches still necessitate extensive training to integrate multimodal inputs and tailor the output from a captioning style to an AD style. 
In this paper, we introduce an automated AD generation pipeline that harnesses the potent multimodal and instruction-following capacities of GPT-4V(ision). 
Notably, our methodology employs readily available components, eliminating the need for additional training. 
It produces ADs that not only comply with established natural language AD production standards but also maintain contextually consistent character information across frames, courtesy of a tracking-based character recognition module. 
A thorough analysis on the MAD dataset reveals that our approach achieves a performance on par with learning-based methods in automated AD production, as substantiated by a CIDEr score of 20.5.

\end{abstract}    
\section{Introduction}
\label{sec:intro}

\begin{figure*}
  \centering	
  \includegraphics[width=1\linewidth]{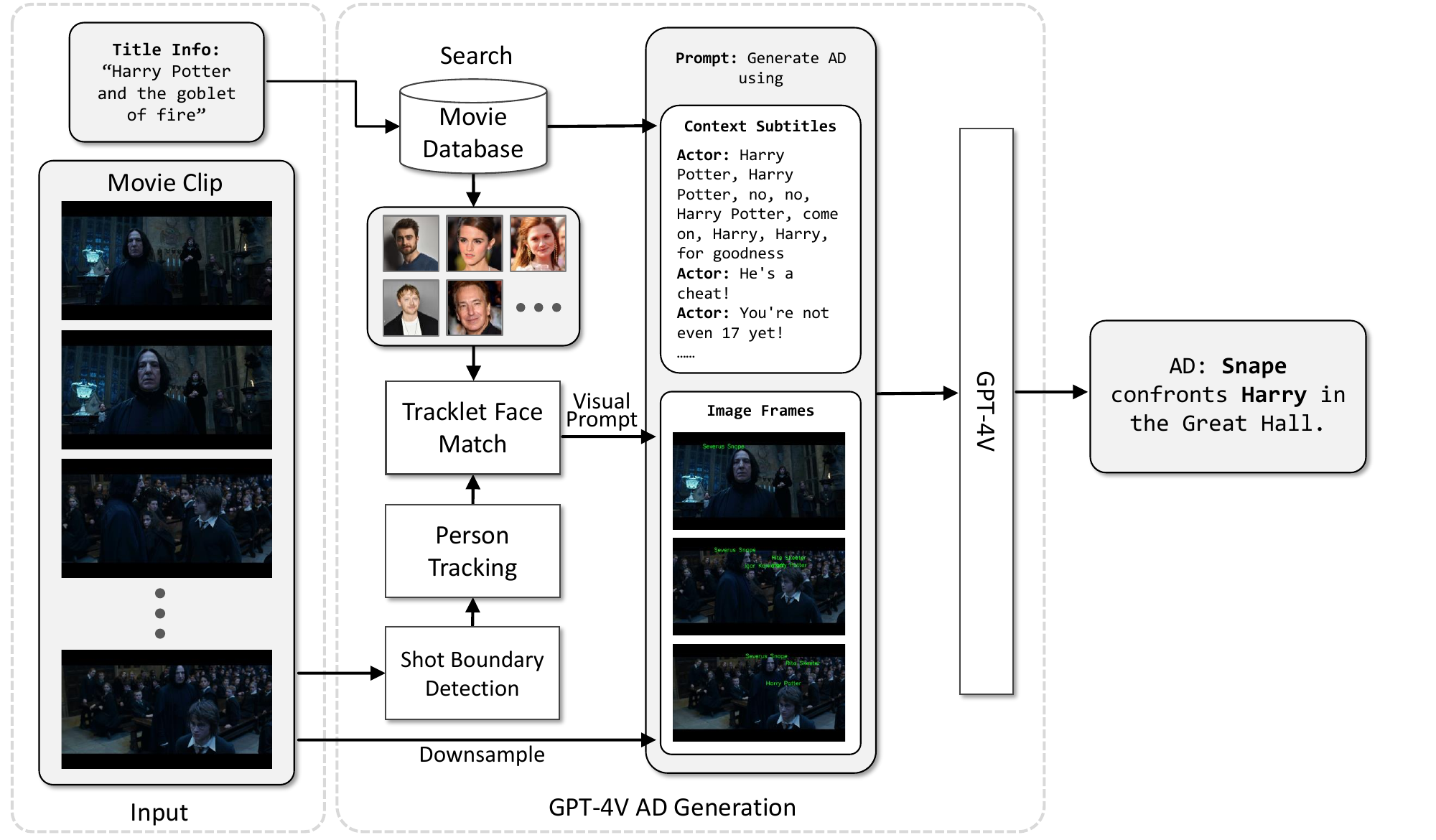}
  \caption{Overview of the GPT-4V based automated AD generation pipeline.}
  \label{fig:overview}
\end{figure*}

The advent of Audio Description (AD) represents a significant leap forward in making video content more accessible and inclusive. 
AD offers a spoken narrative of crucial visual elements within a video that are not captured by the original audio track. 
This innovation holds particular importance for individuals with visual impairments, as it enables them to fully engage with visual media. 
However, creating accurate AD is resource-intensive, requiring specialized expertise, equipment, and significant time investment. 
Automating the production of AD enhances the accessibility of videos for individuals with visual impairments and also broadens the scope of "eyes-free" viewing for the general audience in the realm of popular online videos.

AD is a service that adheres to diverse and well-established production protocols, differing significantly from conventional image captioning tasks by emphasizing storytelling over mere description. 
To generate AD sentences that meet these standards, traditional methods train language models on existing AD datasets to encapsulate the characteristic AD style. 
Another notable challenge in automating AD production is generating sentences of appropriate length that seamlessly fit into the varying temporal gaps within actor dialogue. 
Language models lacking specific instruction-following capabilities often struggle to produce sentences that comply with length constraints. 
However, the advent of Large Language Models (LLMs), particularly GPT-4V(ision), showcases a robust capacity for adhering to instructions. 
This capability enables the generation of more precise AD content by directly feeding the model with AD production guidelines and desired sentence lengths as natural language instruction prompts. 
GPT-4V(ision) stands out as the latest LLM iteration, uniquely supporting both image and text inputs to produce textual outputs. Its advanced multimodal capabilities hold promising potential for establishing a practical automated AD generation framework.


In this paper, we introduce an automated pipeline that utilizes GPT-4V(ision) for the generation of accurate AD for videos. 
Our approach processes a movie clip and its title information to produce AD content. 
We provide an overview of this pipeline, depicted in Fig.~\ref{fig:overview}. 
Our methodology harnesses the multimodal capabilities of GPT-4V, integrating visual cues from video frames with textual context, such as previous subtitles, to generate AD content. 
By allowing the input of AD production guidelines and preferred output sentence lengths as natural language prompts, our system adeptly generates AD of suitable length tailored to speech gaps and can swiftly adapt to various video categories. 
Furthermore, our pipeline incorporates a tracking-based character recognition module, which employs temporal data to deliver consistent character information across frames without necessitating additional training, thus ensuring uniform performance on new video content.

In summary, our contributions are twofold:
\begin{itemize}
    \setlength{\itemindent}{1em}
    \item We propose an automated pipeline that leverages natural language AD production guidelines for the generation of precise AD content.
    \item We introduce a tracking-based character recognition module that supplies GPT-4V with temporally consistent character information.
\end{itemize}

To validate our approach, we conducted extensive experiments on the MAD~\cite{Soldan_2022_CVPR} dataset, with the results setting a new benchmark for state-of-the-art performance.
\section{Related Work}
\label{sec:relatedwork}

\subsection{Large Multimodal Model}

Large multimodal models (LMMs) are a recent trend in artificial intelligence that aims to combine different types of data, such as text, images, audio, and video, to achieve more general and robust intelligence. 
One of the most prominent examples of LMMs is GPT-4V, a model that extends the large language model GPT-4 with vision capabilities. 
GPT-4V can process arbitrarily interleaved multimodal inputs and generate coherent and diverse outputs, such as descriptions, captions, stories, questions, and answers. 
Illustrating its adaptability, GPT-4V has demonstrated outstanding performance across a spectrum of applications, such as visual question answering~\cite{li2023comprehensive}, video interpretation~\cite{lin2023mm}, and clinical diagnostics~\cite{li2023comprehensive, wu2023can}.


\subsection{Audio Description Generation}

Han et al.\cite{han2023autoad1, han2023autoad2} introduced an innovative approach to AD generation by adapting existing language models, such as GPT-2, to handle multimodal information. 
Their methodology involves training adaptation layers using existing AD datasets to imbibe the distinctive linguistic nuances of AD narratives. 
Moreover, the efficacy of their learning-based character recognition module is contingent upon the availability of specialized training datasets.
In contrast, our methodology harnesses natural language instructions to guide the GPT-4V model in producing ADs that adhere to established production guidelines.d 
The proposed method employs a tracking-based character recognition module that operates without the need for training, thus guaranteeing consistent performance across new video content.

MM-VID\cite{lin2023mm} marked a pioneering effort in utilizing the GPT-4V model for AD generation through a unique two-stage pipeline. 
In this process, GPT-4V first synthesizes condensed frame captions, which GPT-4 then refines into the final AD output. 
A critical observation is that their approach lacks an explicit process for character recognition.
Our methodology sets itself apart with a streamlined one-stage process complemented by a dedicated character recognition module. 
This module provides intricate details about characters in the film, enabling the generation of richer and more contextually aligned AD narratives.

\section{Methodology}
\label{sec:method}

Our method takes the video frames of a movie clip and its title information as input to output AD. 
We leverage the visual clues in the video frames, character information and the story context to prompt GPT-4V for the generation as shown in Fig.~\ref{fig:overview}.

\subsection{Character Recognition}

In our approach, we harness the temporal dynamics within videos to execute tracklet-level face recognition, aiming to accurately identify characters from a predefined cast list. 
Initially, upon processing a movie clip's frame sequence, our algorithm employs shot boundary detection techniques to partition the frames into subsequences. 
This partitioning is based on the continuity of camera movement, ensuring that abrupt changes in viewpoint are segmented into distinct subsequences. 
This step is crucial for facilitating the reliable tracking of characters, thereby yielding tracklets that maintain consistent identities throughout.


For each video subsequence, we implement multiple-person tracking to generate person tracklets. 
Tracklets characterized by low confidence scores or brief durations are systematically excluded to eliminate background characters or those not central to the narrative.

To gather comprehensive information about the movie's cast, we query movie databases such as IMDb. 
This search retrieves not only the cast list but also cast profile images and, when available, subtitles. 
It is noteworthy that cast profile images may not always correspond temporally to the period of the movie's production. 
To counteract discrepancies arising from age differences or variations in facial makeup, our method, akin to the strategy outlined in~\cite{han2023autoad2}, involves the collection of exemplar faces of cast members directly from the movie footage. 
Specifically, a face recognition model, denoted as $\mathcal{F}(\cdot)$, processes the cast profile images to produce cast face embeddings. 
This model is subsequently applied to all faces detected within the person tracklets to generate query face embeddings. 
For each cast member, we select the embeddings of the top $K$ matched faces from the query set as augmented embeddings for that cast member's face.


To identify the characters, we compare the face embeddings from each person tracklet against both the original and augmented face embeddings for each cast member. 
We compute an average distance between the tracklet embeddings and those of the cast members. 
The name of the cast member whose embeddings exhibit the shortest average distance to the tracklet and falls below a predefined threshold $\tau$ is assigned as the identity of the character in the tracklet. 
The specifics of this tracklet face matching process are detailed in Alg.~\ref{alg:tracklet_face}. 
Additionally, we explore and evaluate alternative character recognition techniques in Sec.~\ref{sec:tracking}, contributing to a comprehensive understanding of the efficacy of our proposed method.


\begin{algorithm} [!t]
	\small
	\caption{\small Tracklet Face Match}
	\begin{algorithmic}[1]
		\STATE \textbf{Input}: Movie frames $\mathbf{I}$, face images of all casts $\mathbf{I}_{c}$, and person tracklets without name $\mathbf{T}$. 
		\STATE \textbf{Output}: Person tracklets with names $\mathbf{\hat{T}}$.
        \STATE Extract original cast face embedding $\mathbf{E}^{org} \leftarrow \mathcal{F}(\mathbf{I}_{c})$.
        \STATE Extract query face embedding $\mathbf{E}^q \leftarrow \mathcal{F}(\mathsf{crop}(\mathbf{I}, \mathbf{T}))$.
        \FOR {$E_i^{org} \leftarrow E_1^{org},\dots,E_M^{org} $}
            \STATE $\bm{d} \leftarrow \mathsf{dist}(E_i^{org}, \mathbf{E}^q)$ 
            \STATE $\mathbf{E}_i^{aug} \leftarrow \mathbf{E}^q[\mathsf{topk\_index}(-\bm{d}, K)]$ 
        \ENDFOR
        \STATE Cast face embeddings $\mathbf{E}^{c} \leftarrow \mathbf{E}^{org} + \{{E}_1^{aug}, \dots, {E}_M^{aug}\}$
		\FOR {$T_j \leftarrow T_1,\dots,T_{N} $} 
            \STATEx \(\triangleright\) $F_1$ and $F_L$ are the first and last frame \# of $T_j$ 
            \STATE $\mathbf{E}^q_{T_j} \leftarrow \{E^q_{T_j,f=F_1},\dots, E^q_{T_j,f=F_L}\}$
            \FOR {$\mathbf{E}_i^{c} \leftarrow \mathbf{E}_1^{c},\dots,\mathbf{E}_N^{c}$}
                \STATE $D \leftarrow \mathsf{dist}(\mathbf{E}_i^{c}, \mathbf{E}^q_{T_j})$
                \STATE $d_i \leftarrow \mathsf{avg}(\mathsf{flatten}(D))$
            \ENDFOR
            \STATE $\bm{d} \leftarrow \{d_1, \dots, d_M\}$
            \STATEx \(\triangleright\) $\mathsf{argmin}(\cdot)$ returns the index of the smallest element 
            \STATE $i_{\mathrm{min}} \leftarrow \mathsf{argmin}(\bm{d})$
            \IF {$d_{i_\mathrm{min}} < \tau$}
                \STATE $\hat{T}_j \leftarrow \{T_j, i_\mathrm{min}\}$
            \ENDIF
		\ENDFOR
        \STATE $\mathbf{\hat{T}} \leftarrow \{\hat{T}_1, \dots, \hat{T}_N\}$
	\end{algorithmic}
	\label{alg:tracklet_face}
\end{algorithm}

\subsection{Audio Description Generation}

The generation of Audio Descriptions (AD) is a complex process that involves the integration of visual cues, character information, and story context, as illustrated in Fig.\ref{fig:overview}. 
Utilizing GPT-4V, a state-of-the-art language generation model with a vision backbone, enables our system to process both textual and visual inputs, making it ideally suited for our AD generation task.


For visual input, we select 10 image frames in temporal order from each movie clip. 
These frames are processed by GPT-4V, which is instructed to analyze the details within individual frames and to understand actions occurring over multiple frames, effectively interpreting the sequence as a video narrative. 
To incorporate character information, we employ visual prompts by overlaying each image frame with the names of the characters and their corresponding bounding boxes. 
This approach provides GPT-4V with spatial information about the characters, enhancing its ability to generate contextually rich AD. 
The effectiveness of various visual prompts for this task is further examined in Sec.\ref{sec:vprompt}.

Building on the insights from~\cite{han2023autoad1}, we recognize the significance of integrating story context into the AD production process. 
To this end, for each movie clip, we compile any available subtitles within the preceding $T$ AD into the text prompts. 
This ensures the continuity and relevance of the generated AD within the broader narrative context. 
The impact of different choices of $T$ on the quality of the final AD is discussed in Sec.\ref{sec:context}.

Combining the prepared image frames with character annotations and the textual context, along with specific task instructions, we feed this comprehensive input into GPT-4V for AD generation. 
It is imperative to note that the linguistic style required for AD differs significantly from that of typical image captioning tasks. 
AD emphasizes narrative storytelling, whereas image captioning focuses more on describing visuals with even emphasis. 
Our experiments reveal that GPT-4V possesses inherent knowledge about AD and its characteristics, allowing our prompts to simply specify the desired output as AD rather than detailed captioning, thereby conserving token usage.

A crucial aspect of effective AD is ensuring that the length of generated AD is appropriately tailored to fit within the gaps between dialogue or subtitles. 
GPT-4V's advanced instruction-following capabilities allow us to directly control the length of every output AD by specifying the desired word count in the task prompts. 
The influence of different task prompts on AD generation, particularly with respect to controlling AD length, is explored and evaluated in Sec.\ref{sec:adlen}.
\section{Experiments}
\label{sec:experiments}

Our methodology is evaluated using the MAD dataset, as introduced in~\cite{Soldan_2022_CVPR}. 
The MAD dataset is a rich collection comprising over 264,000 audio descriptions sourced from 488 movies. 
For our analysis, we leverage the evaluation subset of this dataset, which includes 10 carefully selected movies, to benchmark our proposed method and conduct comparisons with existing approaches. 
Given the considerable inference time associated with the LLM, we strategically sample approximately 400 movie clips from this subset for all ablation assessments. 
The performance on the complete evaluation subset is reported when compared with existing approaches.


\subsection{Implementation Details}

In our implementation, we utilize a simplified version of the multiple-person tracker presented by~\cite{chu2023transmot} to generate person tracklets, capturing all characters appearing in the input movie clip.
Initial processing involves employing TransNetV2, as described by~\cite{souvcek2020transnet}, for detecting and segmenting clips that contain multiple shots. 
After tracklet generation, we extract square patches around each person from the frames, ensuring these patches include adequate visual context. 
Face detection within these person patches is accomplished using the YOLOv7 model~\cite{yolo7}, enabling us to crop, and align face patches to a standard size of $112 \times 112$ pixels. 
For the purpose of face recognition, we harness the capabilities of AdaFace~\cite{kim2022adaface} with an R100 backbone, trained on the expansive WebFace12M~\cite{zhu2021webface260m} dataset, to extract 512-dimensional face feature vectors. 
For visual inputs concerning each clip, we uniformly sample 10 frames. 
The AD generation phase is powered by the Azure OpenAI GPT-4V deployment.


\vspace{1ex}\noindent\textbf{Evaluation Metrics}
We follow the protocol of other work to adopt classic text generation metrics to report AD generation performance, \eg the classic captioning metrics to compare the generated AD to the ground-truth AD, namely, ROUGE-L~\cite{lin2004rouge} and CIDEr~\cite{vedantam2015cider}.

\subsection{Tracking-based Character Recognition}
\label{sec:tracking}

Our investigation begins with an assessment of the character recognition capabilities across various implemented approaches. 
In this analysis, we evaluate clip-wise character recognition outcomes with their respective ground truths and compute the overall recall and precision across all clips, with results detailed in Tab.~\ref{table:cr}. 
It's pertinent to note that the MAD dataset lacks specific annotations for character recognition. 
To address this, we employ GPT-4 for named entity recognition against a predefined cast list on the ground-truth AD to generate character annotations for each clip. 
Given that not every character is mentioned in the AD, our generated annotations are inherently imperfect and primarily serve to compare the efficacy of different configurations.

\begin{figure}
  \centering	
  \includegraphics[width=1\linewidth]{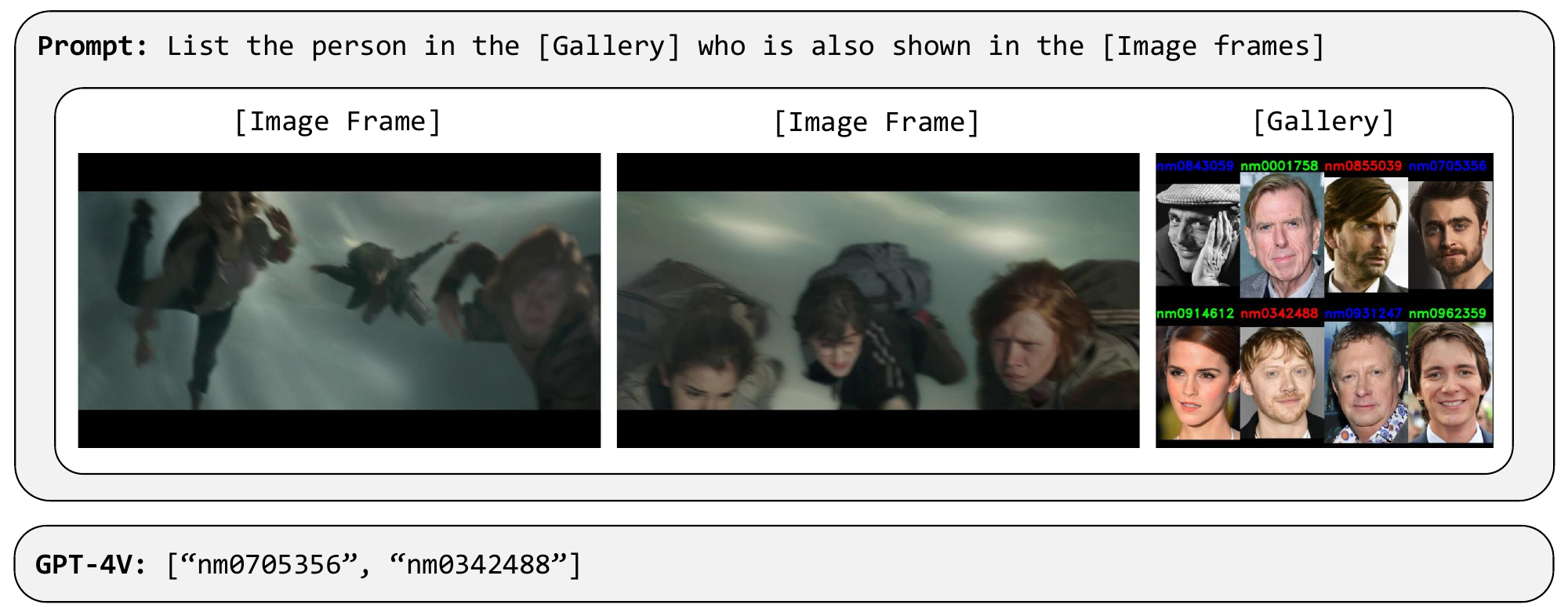}
  \caption{Use GPT-4V for character recognition.}
  \label{fig:gptvcr}
\end{figure}

Given GPT-4V's robust vision capabilities, we establish a GPT-4V based character recognition methodology as our baseline. 
As depicted in Fig.~\ref{fig:gptvcr}, this process involves prompting GPT-4V with nine image frames showing the characters for recognition, supplemented by an additional gallery image showcasing all cast members' faces. 
To mitigate the influence of GPT-4V's pre-existing movie knowledge, we reference the cast using their IMDb IDs rather than their real or character names. GPT-4V is then instructed to identify all observable persons' IDs. 
It's important to highlight that the Azure OpenAI GPT-4V deployment, by default, applies face blurring to user-submitted images depicting humans. 
This precaution significantly impairs recognition accuracy, with GPT-4V's recall and precision approximating random guesses when face blurring is enabled.
Disabling face blurring markedly enhances performance; however, GPT-4V still falls behind in precision compared with a dedicated face recognition model marked as "face recognition only" in Tab.~\ref{table:cr}. 
This discrepancy is partly due to the cast gallery images not always matching the age period of the characters' portrayal in the films. 
Consequently, GPT-4V does not outperform a face recognition model trained with corresponding data in this task.


In our tracking-based character recognition approach, we contrast its performance against scenarios solely utilizing face recognition without tracking.
Additionally, we explore the impact of incorporating exemplar faces. 
By integrating tracking with face recognition, our method achieves a significant improvement in recall, from 0.471 to 0.672, with a minimal decrease in precision of only 0.025. 
Incorporating exemplar faces further enhances recall by 0.037, albeit at a slight cost to precision of 0.004. 
When compared with the existing methodology, where~\cite{han2023autoad2} developed a transformer-based recognition module trained on the MovieNet~\cite{huang2020movienet} dataset, achieving average recall and precision of 0.83 and 0.75 respectively on four selected MAD-eval movies, our proposed tracking-based approach demonstrates a performance disparity in terms of recall and precision. 
However, our method benefits from not requiring additional training, ensuring its performance remains unaffected by the availability of training data and thus offering enhanced generalizability to newly encountered videos.


\begin{table}
	\begin{center}
		\begin{tabular}{c@{\hskip 3.1mm}|@{\hskip 3.5mm}c@{\hskip 3.5mm}c@{\hskip 3.5mm}}
        \hline\thickhline
		Configuration  & Recall & Precision    \\
		\hline
        GPT-4V \textit{w} face blur & 0.342 & 0.276 \\
        GPT-4V \textit{w/o} face blur & 0.468 & 0.518 \\
        \hline
        face recognition only & 0.471 & 0.788 \\
        \textit{w/o} exemplar & 0.672 & 0.763\\
	    ours & 0.709 & 0.759 \\
		\hline\thickhline
		\end{tabular}
	\end{center}
	\vspace{-1mm}
	\caption{Ablations on different character recognition methods and configurations. }
    \label{table:cr}
\end{table}

\subsection{Visual Prompting}
\label{sec:vprompt}

\begin{figure}
	\small
	\centering
	\begin{tabular}{@{\hspace{0.3mm}}c@{\hspace{0.5mm}}c@{\hspace{0.5mm}}}%
  \includegraphics[width=0.49\linewidth]{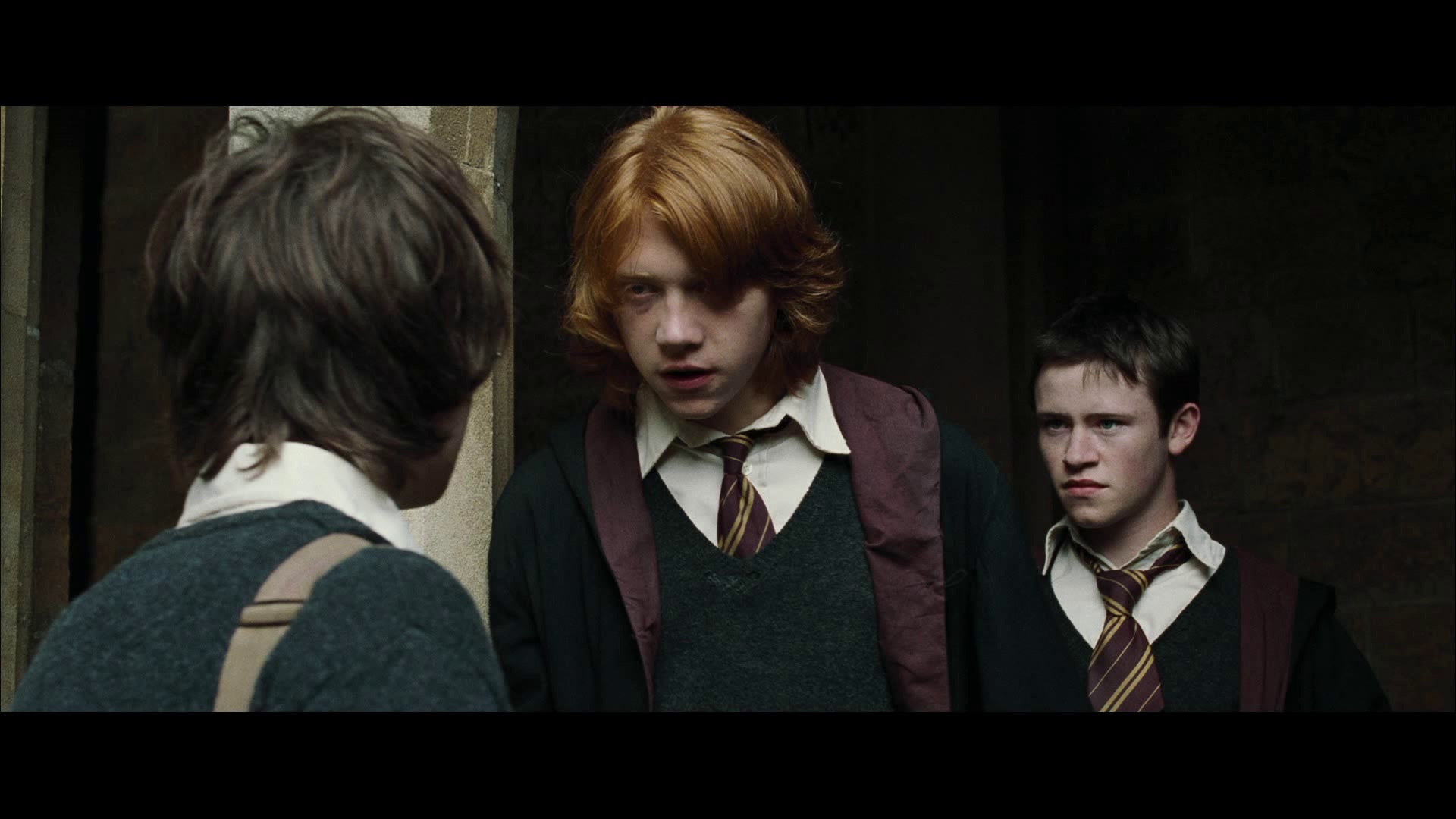} & 
  \includegraphics[width=0.49\linewidth]{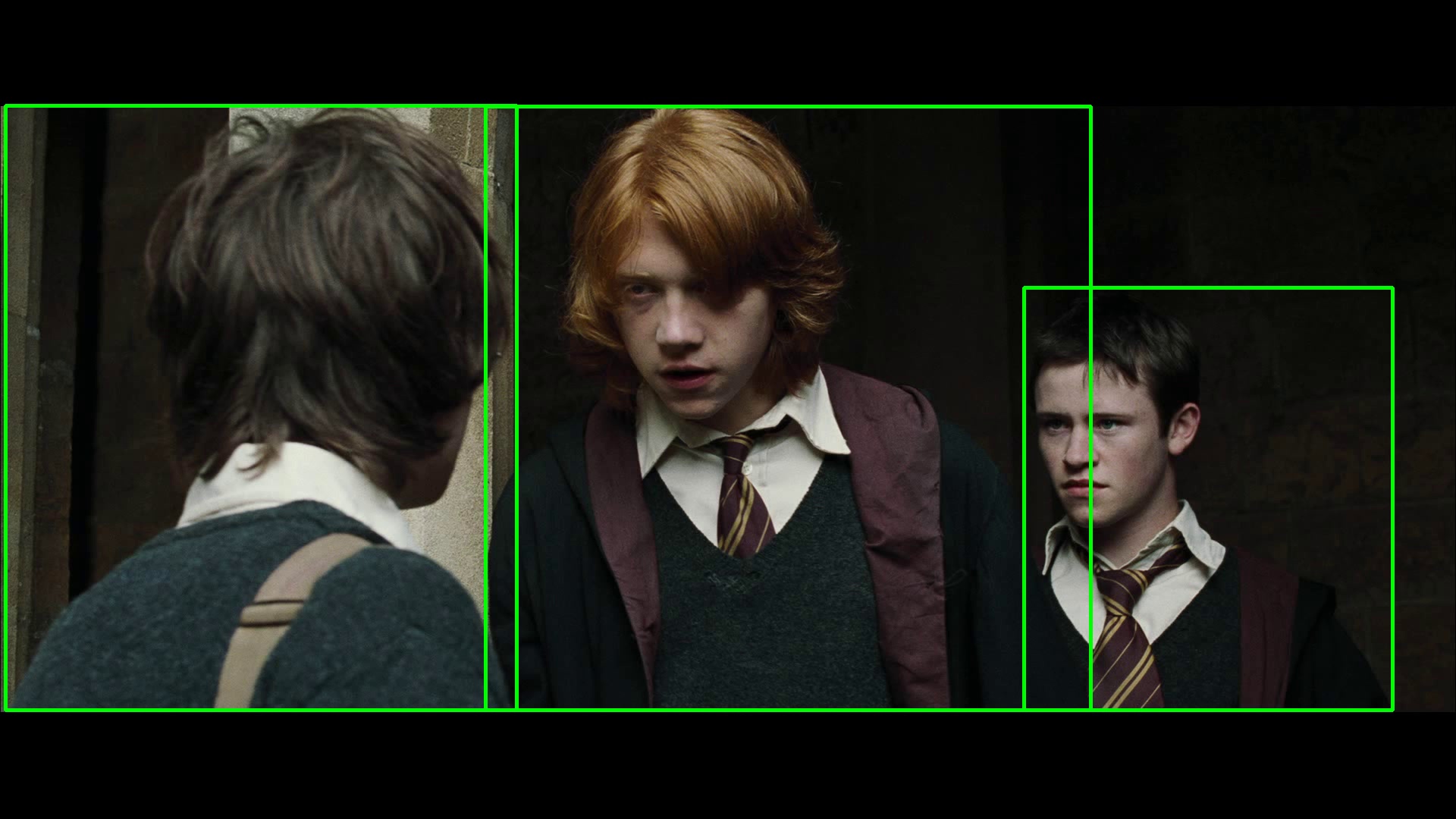} \\
  original frame & bounding box only\\
  \end{tabular}\\

		 \begin{tabular}
		{@{\hspace{0.3mm}}c@{\hspace{0.5mm}}c@{\hspace{0.4mm}}}%
		\includegraphics[width=0.49\linewidth]{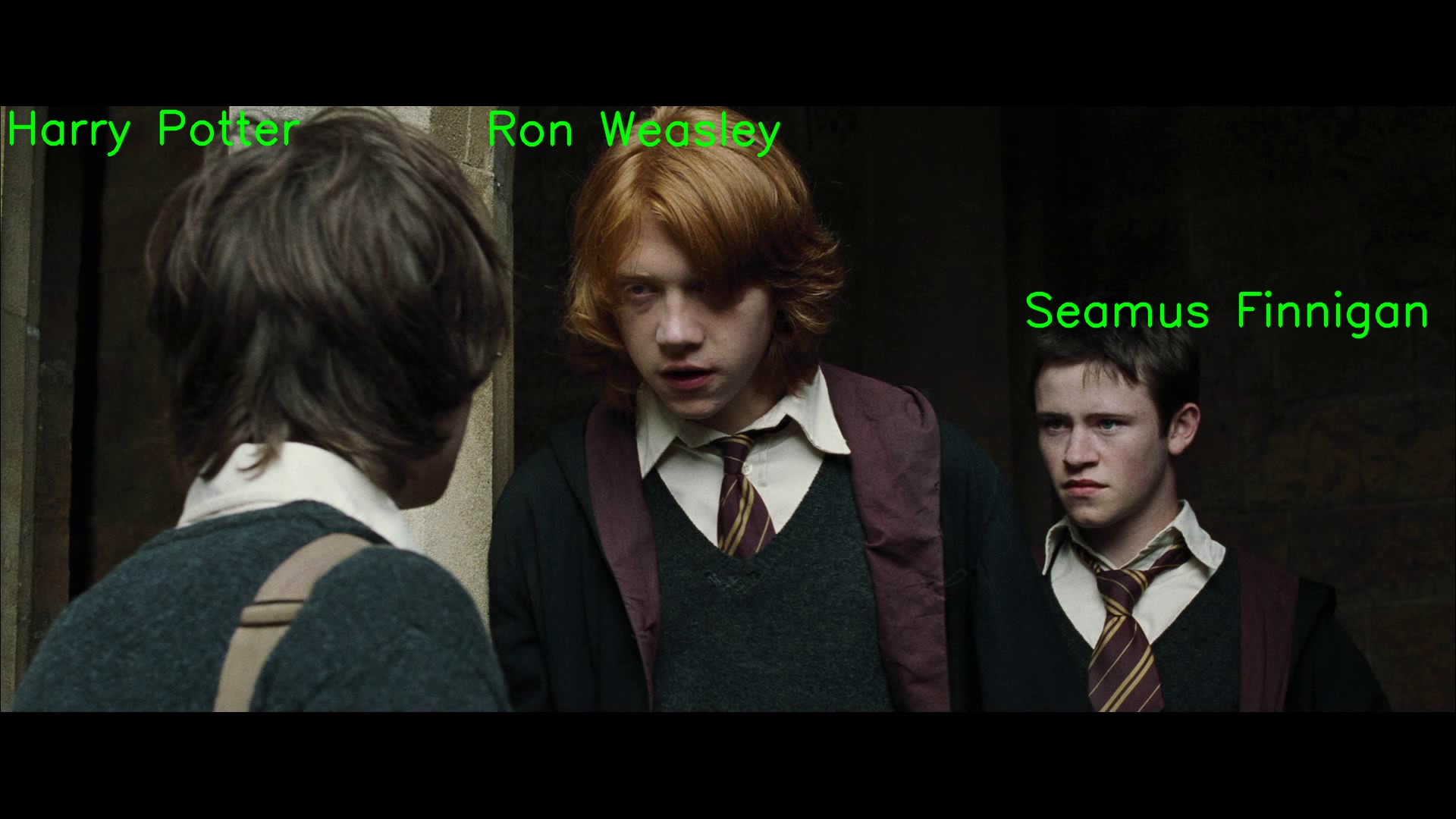}   & 
		\includegraphics[width=0.49\linewidth]{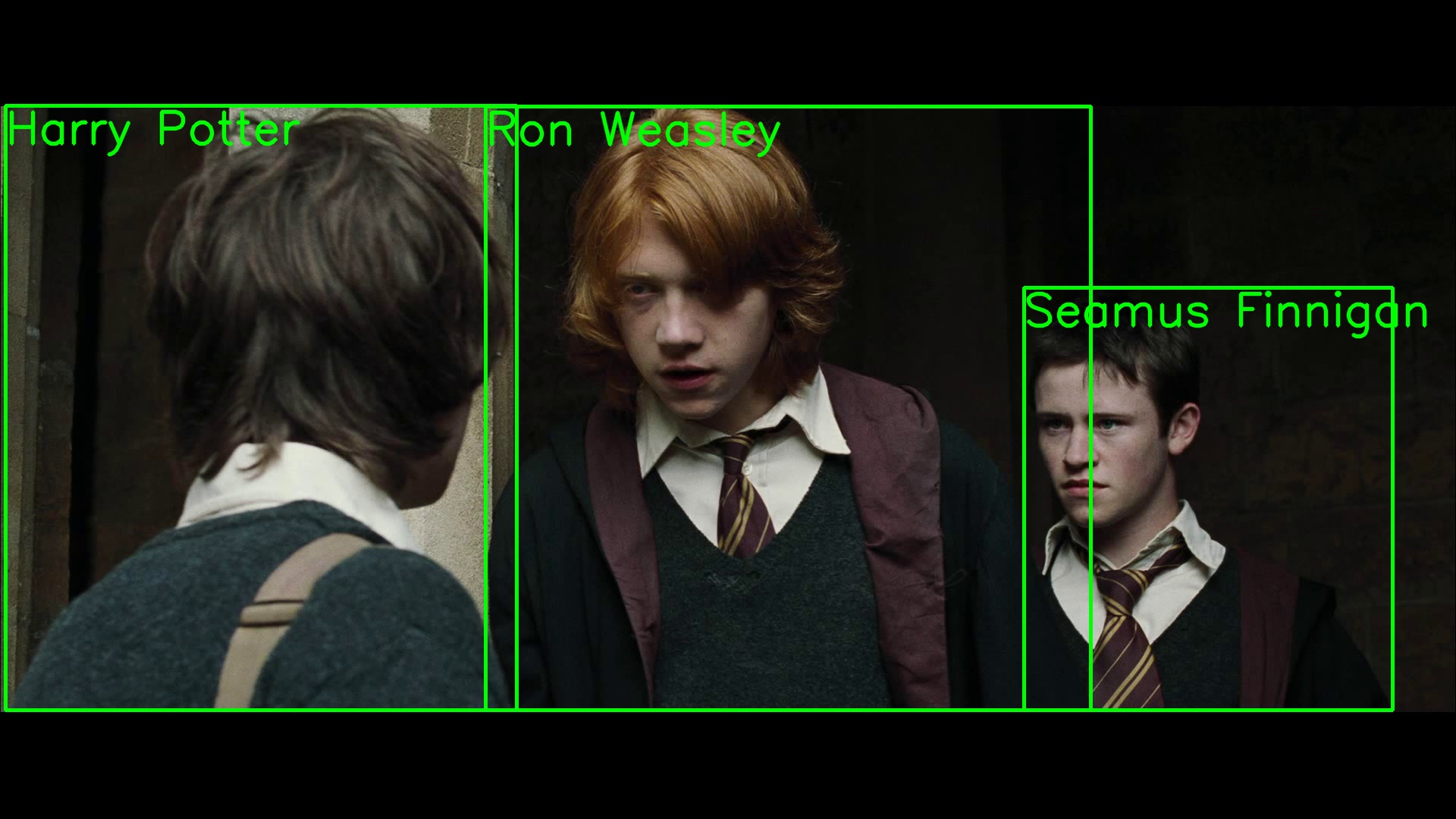}  \\
          name only & bounding box \& name\\
		\end{tabular}
	
		\vspace{3mm}
	\caption{Illustration of different visual prompting to add character information.}
	\label{fig:visprompt}
\end{figure}

In this section, we explore the impact of various visual prompts that incorporate character information into image frames for the generation of AD by GPT-4V. 
As depicted in Fig.~\ref{fig:visprompt}, character information is visually represented by overlaying green bounding boxes and accompanying name tags on the image frames. 
We assess the quality of the generated AD under different settings of visual prompts, with the findings summarized in Tab.~\ref{table:vp}.

When visual prompts consist solely of character bounding boxes, the resulting CIDEr scores decreases 1.8 points compared to those observed in scenarios devoid of any visual prompting. 
This outcome indicates that the inclusion of bounding boxes may obscure pertinent visual cues or detracts from GPT-4V's ability to generate accurate AD. 
Incorporating only the names of characters directly onto the image frames significantly enhances the quality of the generated AD, evidenced by a notable increase in the CIDEr score (from 18.8 to 23.9) and ROUGE-L (from 9.6 to 13.6). 
Further adding bounding boxes to the character names decreases the CIDEr score by 0.9. 
Based on these results, we opted to utilize character name text only in subsequent experiments.



\begin{figure}
  \centering	
  \includegraphics[width=1\linewidth]{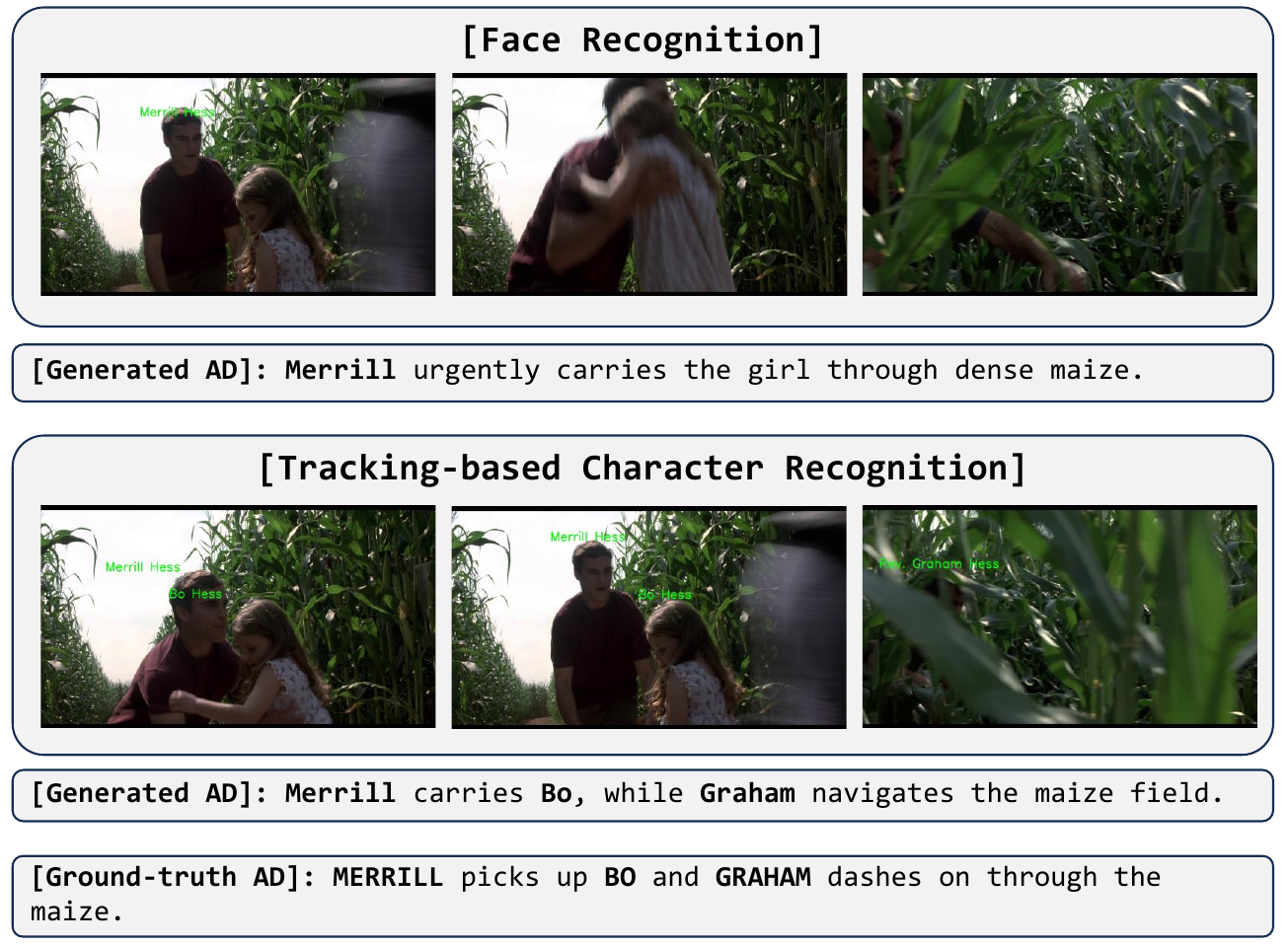}
  \caption{Generated AD with different character recognition methods.}
  \label{fig:tracking}
\end{figure}

The last entry in Tab.~\ref{table:vp} demonstrates the comprehensive effectiveness of our tracking-based character recognition module by contrasting the AD quality against that produced using only a standard face recognition approach. 
Employing a conventional face recognition technique, GPT-4V generates AD with a notably lower CIDEr score (\eg, 18.8 vs. 23.9). 
An illustrative comparison presented in Fig.~\ref{fig:tracking} reveals that, without the aid of tracking, the system struggles to deduce character names when faces are either occluded or not facing the camera, resulting in the omission of two characters' names in the generated AD. 


\begin{table}
	\begin{center}
		\begin{tabular}{c@{\hskip 3.5mm}|@{\hskip 3.5mm}c@{\hskip 3.5mm}c@{\hskip 3.5mm}}
        \hline\thickhline
		Configuration  & ROUGE-L & CIDEr    \\
		\hline
        \textit{w/o} visual prompting & 9.6 & 18.8 \\
        bounding box only & 9.9  & 17.0  \\
        name only & 13.6 & 23.9  \\
	    bounding box \& name & 12.4 & 23.0  \\
        \hline
        face recognition only & 10.8 & 18.8 \\
		\hline\thickhline
		\end{tabular}
	\end{center}
	\vspace{-1mm}
	\caption{Ablations on different visual prompting for adding character names. }
    \label{table:vp}
\end{table}

\subsection{Textual Context}
\label{sec:context}
The incorporation of textual context significantly enhances the quality of the generated AD, as demonstrated in~\cite{han2023autoad1}. 
We investigate the impact of varying the number of textual context, \eg subtitles and previous AD, included in the prompt on the generated AD quality, with our findings presented in Tab.~\ref{table:context}. 
To quantify the context length, we follow the definition in~\cite{han2023autoad1} to treat the subtitles between the $T$-th AD and the current AD timestamp as context length $T$.
Notably, the inclusion of 100 context subtitles markedly boosts the CIDEr score from 22.2 to 23.9 and the ROUGE-L score from 12.8 to 13.5, in comparison to scenarios without any context subtitles. 
This substantial improvement underscores the value of leveraging subtitles to generate context-coherent AD. 
However, augmenting the inputs with more than 100 context subtitles yields diminishing returns in performance enhancement, as illustrated by the results obtained for 200 contexts.

The role of previous AD in enhancing AD performance is also evident. 
Referring to the last row of Tab.~\ref{table:context}, the inclusion of previous AD leads to a decrease in the CIDEr score by 1.6 points, from 23.9 to 22.3. 
This decrease in performance could be partially attributed to our asynchronous inference design. 
In order to process extensive videos such as movies, we have structured the pipeline to infer input clips in parallel. 
To incorporate previous AD as textual context, the pipeline must be executed twice. 
During the initial run, the AD is generated solely based on context subtitles. 
In the subsequent run, the AD generated from the previous step is incorporated as context AD to produce the final AD. 
This design deviates from the recursive configuration in\cite{han2023autoad1}, and as a result, may not fully exploit the information encapsulated in the context AD.

\begin{table}[!h]
	\begin{center}
		\begin{tabular}{c@{\hskip 3.1mm}|@{\hskip 3.5mm}c@{\hskip 3.5mm}c@{\hskip 3.5mm}}
        \hline\thickhline
        Number of Context $T$ & ROUGE-L & CIDEr    \\
		\hline
        0 & 12.8 & 22.2  \\
        6 & 13.1 & 23.2  \\
        20 & 13.0 & 23.5 \\
	    100 & 13.6 & 23.9 \\
        200 & 13.1 & 20.0 \\
        \hline
        100 \textit{w} context AD & 12.4 & 22.3 \\
		\hline\thickhline
		\end{tabular}
	\end{center}
	\vspace{-1mm}
	\caption{Ablations on the number of context AD and subtitles to include for AD generation.}
    \label{table:context}
\end{table}

\subsection{Prompts for Audio Description Generation}
\label{sec:adlen}

\begin{figure}[!h]
  \centering	
  \includegraphics[width=1\linewidth]{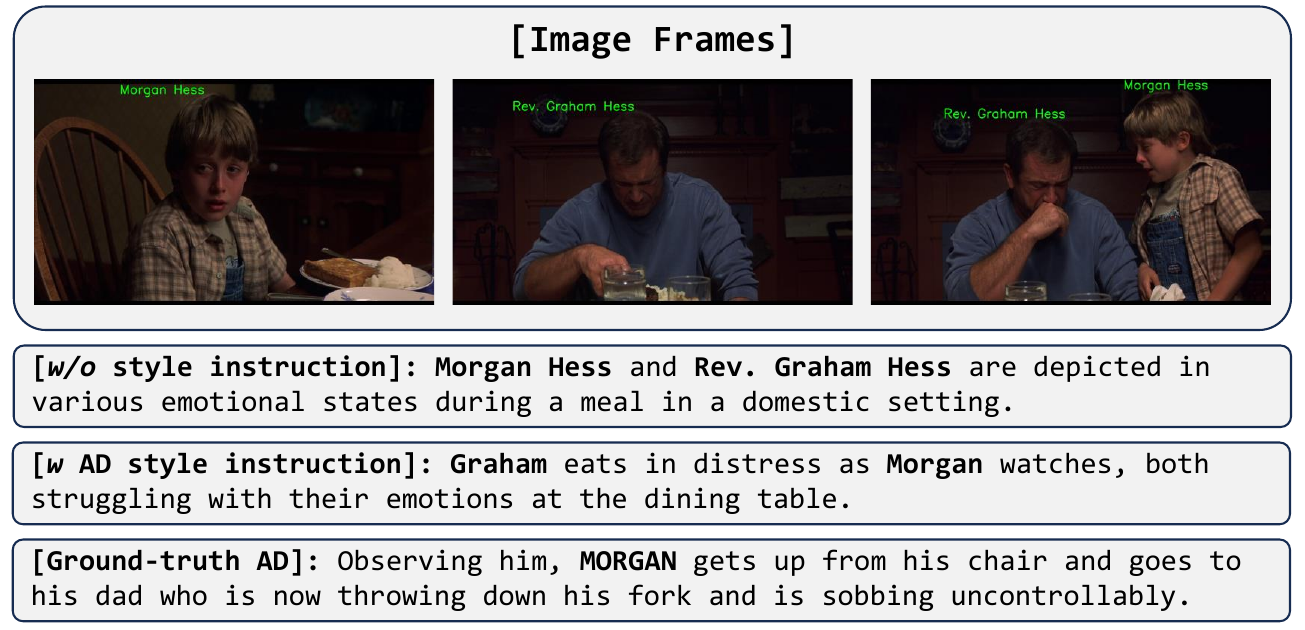}
    \includegraphics[width=1\linewidth]{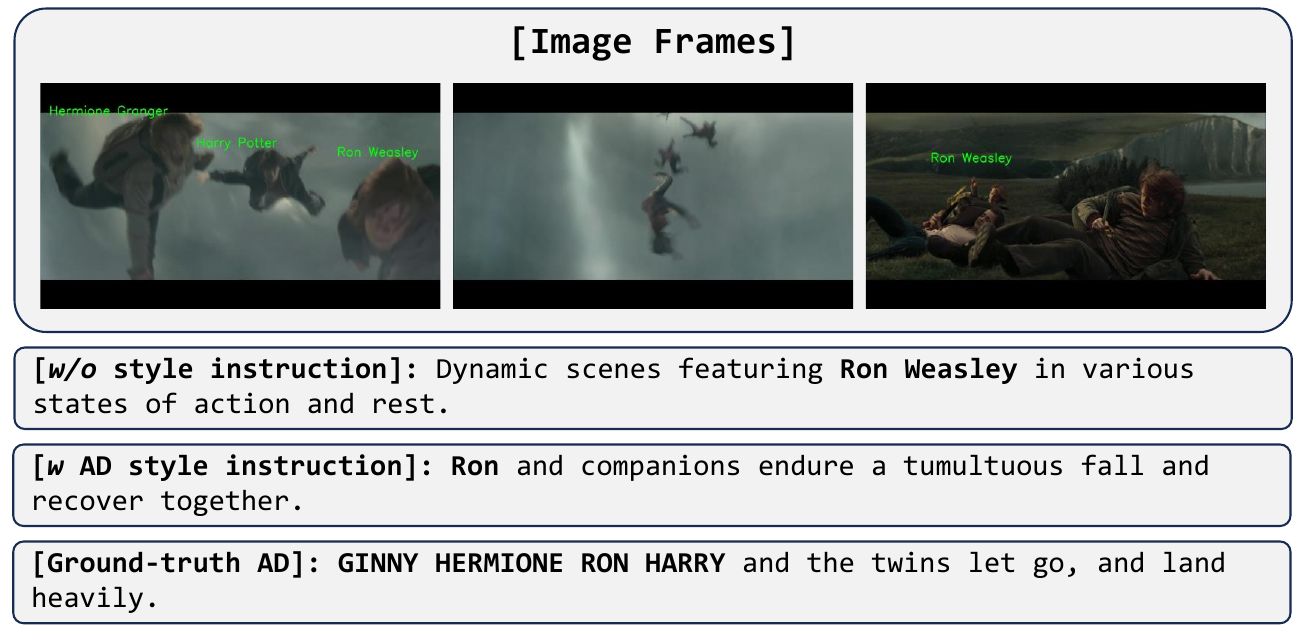}
  \caption{Generated AD with different linguistic styles instructions.}
  \label{fig:adstyle}
\end{figure}

In this section, we delve into the effects of varying task prompts on the generation of AD by comparing the performance impact of explicitly specifying an AD-style output against not doing so. 
The initial portion of Tab.~\ref{table:adlen} presents a comparison between using a task prompt that directs GPT-4V to generate AD-style content and a prompt that does not. 
To isolate the influence from textual context, no contextual AD or subtitles were included, preventing GPT-4V from context learning the AD style from prior examples. 
The findings indicate a significant enhancement in the quality of AD generation, with a CIDEr score increase of 6.4 when GPT-4V is instructed to produce AD-style outputs as opposed to standard caption-style sentences. 
Fig.~\ref{fig:adstyle} showcases two examples illustrating that, under restricted word count conditions, a captioning-style output tends to enumerate observations in the image sequences, whereas an AD-style output weaves these observations into a narrative more akin to the ground-truth AD.


Another critical aspect of high-quality AD production is ensuring the generated AD appropriately fits within the time gaps of subtitles. 
Without explicit word count constraints, GPT-4V's output tends to be verbose, significantly diverging from the concise style of ground-truth AD. 
The second section of Tab.~\ref{table:adlen} highlights this issue, showing a mere 4.7 CIDEr score for AD generated without word count limitations. 
In contrast to other methods that infer a concise AD style through learning from training data, GPT-4V's robust instruction-following capability allows us to directly specify the desired word count for AD outputs. 
We explore the impact of instructing GPT-4V to generate all AD in fixed word counts, such as 6, 10 and 20 words, with the performance outcomes detailed in Tab.~\ref{table:adlen}. 
The choice of 6 words aligns with the average word count across the AudioVault~\cite{han2023autoad1} dataset, where 80\% of the AD contain 10 words or fewer, and 99\% of the AD do not exceed 20 words. 
Our results demonstrate that, amongst the fixed word counts of 6, 10, and 20, the 10-word prompts exhibit the highest ROUGE-L and CIDEr scores.


Given the variability in the duration of subtitle gaps, optimally, ADs should vary in length to match these intervals. 
In this study, we employ a setting that references the word count of each ground-truth AD to guide GPT-4V in generating AD of corresponding lengths, aiming to illustrate the importance of AD length. 
The potential for automatically estimating AD length based on the temporal gaps between subtitles represents an exciting avenue for future research. 
The results in Tab.~\ref{table:adlen} demonstrate the superior performance of the variant length AD setting, outperforming the 10-word setting by 3.3 points in the CIDEr metric. 


The MM-VID~\cite{lin2023mm} framework employs a two-stage approach for generating AD, where GPT-4V is utilized for generating clip-level descriptions, and GPT-4 is tasked with producing task-specific responses based on the output from GPT-4V. 
Inspired by this methodology, we too investigated a similar two-stage architecture. 
In our exploration, we directed GPT-4V to create detailed textual descriptions for each of the 10 image frames. 
Subsequently, GPT-4 was instructed to summarize GPT-4V’s detailed outputs into coherent AD, applying similar prompts discussed earlier (e.g., context and output word count constraints). 
The outcomes, as reported in the final section of Tab.~\ref{table:adlen}, reveal a inferior performance in comparison to the results from the preceding setup (CIDEr from 23.9 to 21.5, ROUGE-L from 13.6 to 12.3).

This decline in performance could be attributed to the design intent behind MM-VID, which is crafted as a general tool for video understanding, necessitating GPT-4’s superior instruction-following capabilities to manage multiple tasks concurrently. 
In contrast, our framework is specifically tailored for the automatic generation of AD, for which the linguistic capabilities of GPT-4V alone appear to be adequately sufficient. 
Moreover, the streamlined one-stage design may offer the added advantage of preserving subtle visual cues that could potentially be overlooked in the transition from frame-level textual descriptions to final AD output in a two-stage process.


\begin{table}[!h]
	\begin{center}
		\begin{tabular}{c@{\hskip 3.1mm}|@{\hskip 3.5mm}c@{\hskip 3.5mm}c@{\hskip 3.5mm}}
        \hline\thickhline
        Prompts & ROUGE-L & CIDEr    \\
        \hline
        \textit{w/o} output style instruction & 11.1 & 17.5\\
        \textit{w} AD style instruction & 12.6 & 23.9\\
		\hline
        \textit{w/o} output word count limit & 12.1  &  4.7  \\
        output 6 words & 10.4 & 19.0 \\
	    output 10 words & 13.6  & 23.9 \\
        output 20 words & 13.1  & 19.8 \\
        output ground-truth AD length & 12.9 & 27.2  \\
        \hline
        two-stage, GPT-4 summary & 12.3 & 21.5
 \\
		\hline\thickhline
		\end{tabular}
	\end{center}
	\vspace{-1mm}
	\caption{Ablations on different prompts for AD generation.}
    \label{table:adlen}
\end{table}

\subsection{Comparison with the SOTA}

For an equitable comparison with existing methodologies, we assess our approach using the complete evaluation subset of the MAD dataset.
To ensure a fair comparison with alternative approaches, we configured our method to generate AD consisting of a fixed length of 10 words. 
Our approach demonstrates superior performance over AutoAD-II, establishing a new state-of-the-art performance with CIDEr and ROUGE-L scores of 20.5 (\textit{vs} 19.5) and 13.5 (\textit{vs} 13.4), respectively.


\begin{table}[!h]
	\begin{center}
		\begin{tabular}{c@{\hskip 3.1mm}|@{\hskip 3.5mm}c@{\hskip 3.5mm}|c@{\hskip 3.5mm}c@{\hskip 3.5mm}}
        \hline\thickhline
        Methods & context & ROUGE-L & CIDEr    \\
		\hline
        ClipCap~\cite{mokady2021clipcap} & no & 8.5 &  4.4  \\
        AutoAD-I~\cite{han2023autoad1} & no & 10.3 & 12.1  \\
        AutoAD-II~\cite{han2023autoad2} & no & 13.1 & 19.2 \\
        \hline
        AutoAD-I~\cite{han2023autoad1} & AD \& subt.&11.9 & 14.3 \\
        AutoAD-II~\cite{han2023autoad2} & AD \& subt.&13.4  & 19.5 \\
        Ours & subt. &13.5 & 20.5 \\

		\hline\thickhline
		\end{tabular}
	\end{center}
	\vspace{-1mm}
	\caption{Comparison with other methods.}
    \label{table:absperf}
\end{table}

\section{Discussion and Future Work}

The GPT-4V models, trained on extensive collections of visual and linguistic data, represent a significant stride forward in the automatic generation of Audio Descriptions (AD) for cinematic content. 
However, the generation of AD for films within the MAD dataset by GPT-4V may inherently incorporate biases due to the model's exposure to potentially similar content during training. 
To mitigate such biases, it is advisable to evaluate the model's performance using a more contemporary selection of movies, which are less likely to have been part of the model's training corpus.

Another notable limitation of our current approach is the absence of a mechanism for determining appropriate moments within a film to insert AD and estimating the corresponding word count for that AD. 
This deficiency becomes particularly evident when considering the substantial decline in performance (a CIDEr reduction from 27.2 to 23.9) observed when imposing a fixed word count for AD output, as opposed to tailoring the length based on the ground truth AD.
The performance gap shows the importance of AD length and exhibits a promising direction for future works to explore for improving the generated AD quality. 
As an example, one can customize a relatively lightweight language-rewritten model from existing AD data to tailor the over-verbose output from the LLM.


{
    \small
    \bibliographystyle{ieeenat_fullname}
    \bibliography{main}

\begin{thebibliography}{15}
\providecommand{\natexlab}[1]{#1}
\providecommand{\url}[1]{\texttt{#1}}
\expandafter\ifx\csname urlstyle\endcsname\relax
  \providecommand{\doi}[1]{doi: #1}\else
  \providecommand{\doi}{doi: \begingroup \urlstyle{rm}\Url}\fi

\bibitem[yol()]{yolo7}
yolo7-face.
\newblock \url{https://github.com/derronqi/yolov7-face}.

\bibitem[Chu et~al.(2023)Chu, Wang, You, Ling, and Liu]{chu2023transmot}
Peng Chu, Jiang Wang, Quanzeng You, Haibin Ling, and Zicheng Liu.
\newblock Transmot: Spatial-temporal graph transformer for multiple object tracking.
\newblock In \emph{Proceedings of the IEEE/CVF Winter Conference on Applications of Computer Vision}, pages 4870--4880, 2023.

\bibitem[Han et~al.(2023{\natexlab{a}})Han, Bain, Nagrani, Varol, Xie, and Zisserman]{han2023autoad1}
Tengda Han, Max Bain, Arsha Nagrani, G\"ul Varol, Weidi Xie, and Andrew Zisserman.
\newblock {AutoAD}: Movie description in context.
\newblock In \emph{CVPR}, 2023{\natexlab{a}}.

\bibitem[Han et~al.(2023{\natexlab{b}})Han, Bain, Nagrani, Varol, Xie, and Zisserman]{han2023autoad2}
Tengda Han, Max Bain, Arsha Nagrani, G\"ul Varol, Weidi Xie, and Andrew Zisserman.
\newblock {AutoAD II: The Sequel} - who, when, and what in movie audio description.
\newblock In \emph{ICCV}, 2023{\natexlab{b}}.

\bibitem[Huang et~al.(2020)Huang, Xiong, Rao, Wang, and Lin]{huang2020movienet}
Qingqiu Huang, Yu Xiong, Anyi Rao, Jiaze Wang, and Dahua Lin.
\newblock Movienet: A holistic dataset for movie understanding.
\newblock In \emph{Computer Vision--ECCV 2020: 16th European Conference, Glasgow, UK, August 23--28, 2020, Proceedings, Part IV 16}, pages 709--727. Springer, 2020.

\bibitem[Kim et~al.(2022)Kim, Jain, and Liu]{kim2022adaface}
Minchul Kim, Anil~K Jain, and Xiaoming Liu.
\newblock Adaface: Quality adaptive margin for face recognition.
\newblock In \emph{Proceedings of the IEEE/CVF Conference on Computer Vision and Pattern Recognition}, 2022.

\bibitem[Li et~al.(2023)Li, Liu, Wang, Liang, Liu, Wang, Cui, Tu, Wang, and Zhou]{li2023comprehensive}
Yingshu Li, Yunyi Liu, Zhanyu Wang, Xinyu Liang, Lingqiao Liu, Lei Wang, Leyang Cui, Zhaopeng Tu, Longyue Wang, and Luping Zhou.
\newblock A comprehensive study of gpt-4v's multimodal capabilities in medical imaging.
\newblock \emph{medRxiv}, pages 2023--11, 2023.

\bibitem[Lin(2004)]{lin2004rouge}
Chin-Yew Lin.
\newblock Rouge: A package for automatic evaluation of summaries.
\newblock In \emph{Text summarization branches out}, pages 74--81, 2004.

\bibitem[Lin et~al.(2023)Lin, Ahmed, Li, Lin, Azarnasab, Yang, Wang, Liang, Liu, Lu, et~al.]{lin2023mm}
Kevin Lin, Faisal Ahmed, Linjie Li, Chung-Ching Lin, Ehsan Azarnasab, Zhengyuan Yang, Jianfeng Wang, Lin Liang, Zicheng Liu, Yumao Lu, et~al.
\newblock Mm-vid: Advancing video understanding with gpt-4v (ision).
\newblock \emph{arXiv preprint arXiv:2310.19773}, 2023.

\bibitem[Mokady et~al.(2021)Mokady, Hertz, and Bermano]{mokady2021clipcap}
Ron Mokady, Amir Hertz, and Amit~H Bermano.
\newblock Clipcap: Clip prefix for image captioning.
\newblock \emph{arXiv preprint arXiv:2111.09734}, 2021.

\bibitem[Soldan et~al.(2022)Soldan, Pardo, Alc\'azar, Caba, Zhao, Giancola, and Ghanem]{Soldan_2022_CVPR}
Mattia Soldan, Alejandro Pardo, Juan~Le\'on Alc\'azar, Fabian Caba, Chen Zhao, Silvio Giancola, and Bernard Ghanem.
\newblock Mad: A scalable dataset for language grounding in videos from movie audio descriptions.
\newblock In \emph{Proceedings of the IEEE/CVF Conference on Computer Vision and Pattern Recognition (CVPR)}, pages 5026--5035, 2022.

\bibitem[Sou{\v{c}}ek and Loko{\v{c}}(2020)]{souvcek2020transnet}
Tom{\'a}{\v{s}} Sou{\v{c}}ek and Jakub Loko{\v{c}}.
\newblock Transnet v2: An effective deep network architecture for fast shot transition detection.
\newblock \emph{arXiv preprint arXiv:2008.04838}, 2020.

\bibitem[Vedantam et~al.(2015)Vedantam, Lawrence~Zitnick, and Parikh]{vedantam2015cider}
Ramakrishna Vedantam, C Lawrence~Zitnick, and Devi Parikh.
\newblock Cider: Consensus-based image description evaluation.
\newblock In \emph{Proceedings of the IEEE conference on computer vision and pattern recognition}, pages 4566--4575, 2015.

\bibitem[Wu et~al.(2023)Wu, Lei, Zheng, Zhao, Lin, Zhang, Zhou, Zhao, Zhang, Wang, et~al.]{wu2023can}
Chaoyi Wu, Jiayu Lei, Qiaoyu Zheng, Weike Zhao, Weixiong Lin, Xiaoman Zhang, Xiao Zhou, Ziheng Zhao, Ya Zhang, Yanfeng Wang, et~al.
\newblock Can gpt-4v (ision) serve medical applications? case studies on gpt-4v for multimodal medical diagnosis.
\newblock \emph{arXiv preprint arXiv:2310.09909}, 2023.

\bibitem[Zhu et~al.(2021)Zhu, Huang, Deng, Ye, Huang, Chen, Zhu, Yang, Lu, Du, et~al.]{zhu2021webface260m}
Zheng Zhu, Guan Huang, Jiankang Deng, Yun Ye, Junjie Huang, Xinze Chen, Jiagang Zhu, Tian Yang, Jiwen Lu, Dalong Du, et~al.
\newblock Webface260m: A benchmark unveiling the power of million-scale deep face recognition.
\newblock In \emph{Proceedings of the IEEE/CVF Conference on Computer Vision and Pattern Recognition}, pages 10492--10502, 2021.

\end{thebibliography}
}

\section{Appendix}
\label{sec:appendix}
\begin{figure*}
  \centering	
  \includegraphics[width=1\linewidth]{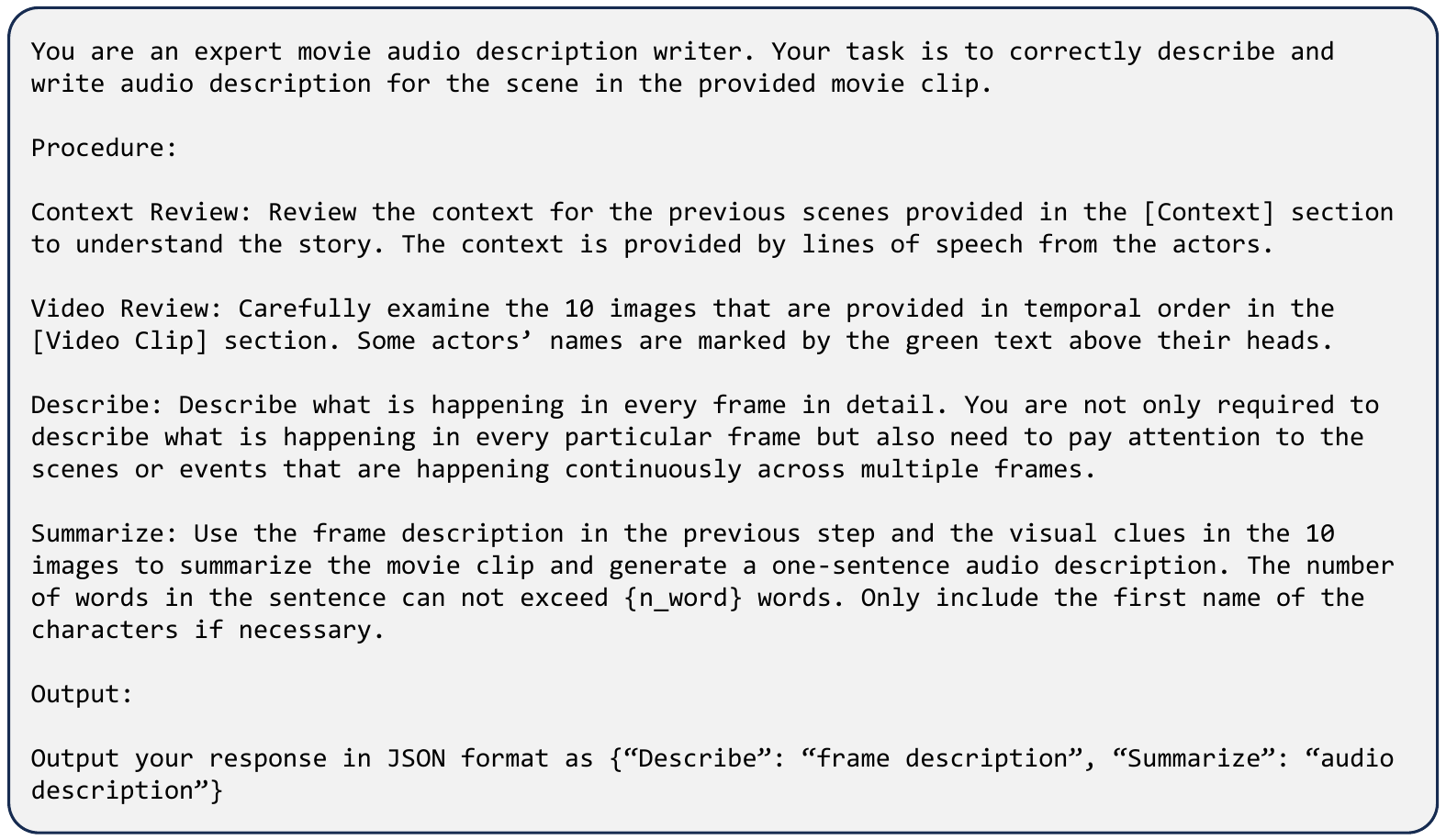}
  \caption{GPT-4V prompt to generate the audio description.}
  \label{fig:prompt}
\end{figure*}

The GPT-4V prompt we used to generate the audio description is listed in Fig.\ref{fig:prompt}.
%

\end{document}